\documentclass[journal]{IEEEtran}

\ifCLASSINFOpdf
\else
   \usepackage[dvips]{graphicx}
\fi
\usepackage{url}

\usepackage{graphicx}
\usepackage{color}
\usepackage{amsmath}
\usepackage{bm}
\usepackage{float}
\usepackage{booktabs}
\usepackage{cite}
\usepackage{makecell} 
\usepackage{amssymb}
\usepackage{caption}
\usepackage{algorithm}
\usepackage{algorithmic}
\usepackage{multirow}
\usepackage{pifont}
\usepackage{hyperref}
\usepackage{subcaption}
\usepackage{hyperref} 

\newcommand{\RN}[1]{\MakeUppercase{\romannumeral #1}}

\begin{document}

\title{Towards Comprehensive Information-theoretic Multi-view Learning}    

\author{Long Shi, \IEEEmembership{Member, IEEE}, Yunshan Ye, Wenjie Wang, Tao Lei, Yu Zhao, Gang Kou, Badong Chen, \IEEEmembership{Senior Member, IEEE}
\thanks{Long Shi, Wenjie Wang and Yu Zhao are with the School of Computing and Artificial Intelligence, and also with Artificial Intelligence and Digital Finance Key Laboratory of Sichuan Province, Southwestern University of Finance and Economics, Chengdu 611130, China (e-mail: shilong@swufe.edu.cn, 224081200036@smail.swufe.edu.cn, zhaoyu@swufe.edu.cn)}
\thanks{Yunshan Ye is with the School of Business Administration, Southwestern University of Finance and Economics, Chengdu 611130, China (e-mail: 1241201z5006@smail.swufe.edu.cn)}
\thanks{Tao Lei is with the Institute of Optics and Electronics, Chinese Academy of Sciences, Chengdu 610209, China (e-mail: taoleiyan@ioe.ac.cn)}
\thanks{Gang Kou is with Xiangjiang Laboratory, Changsha 410205, China. He is also affiliated with the School of Digital Media Engineering and Humanities, Hunan University of Technology and Business, Changsha 410205, China, and the School of Business Administration, Faculty of Business Administration, Southwestern University of Finance and Economics, Chengdu 610074, China.}
\thanks{Badong Chen is with the Institute of Artificial Intelligence and Robotics, Xi’an Jiaotong University, Xi’an 710049, China (e-mail: chenbd@mail.xjtu.edu.cn)}}


\markboth{Journal of \LaTeX\ Class Files}
{Shell \MakeLowercase{\textit{et al.}}: Bare Demo of IEEEtran.cls for IEEE Journals}
\maketitle

\begin{abstract}
Information theory has inspired numerous advancements in multi-view learning. Most multi-view methods incorporating information-theoretic principles rely an assumption called {\it multi-view redundancy} which states that common information between views is necessary and sufficient for down-stream tasks. This assumption emphasizes the importance of common information for prediction, but inherently ignores the potential of unique information in each view that could be predictive to the task. In this paper, we propose a comprehensive information-theoretic multi-view learning framework named CIML, which discards the assumption of {\it multi-view redundancy}. Specifically, CIML considers the potential predictive capabilities of both common and unique information based on information theory. First, the common representation learning maximizes G$\acute{a}$cs-K$\mathaccent"707F o$rner common information to extract shared features and then compresses this information to learn task-relevant representations based on the Information Bottleneck (IB). For unique representation learning, IB is employed to achieve the most compressed unique representation for each view while simultaneously minimizing the mutual information between unique and common representations, as well as among different unique representations. Importantly, we theoretically prove that the learned joint representation is predictively sufficient for the downstream task. Extensive experimental results have demonstrated the superiority of our model over several state-of-art methods. The code is released on \href{https://github.com/ahahaaho996/CIML}{CIML}.
\end{abstract}



\section{Introduction}

\IEEEPARstart{I}{n} real-world applications, data collection frequently occurs from multiple sources, which promotes the development of multi-view learning \cite{zhao2017multi}. For instance, driver monitoring systems analyze synchronized video streams, driver body posture, and facial expressions to holistically assess states like distraction or fatigue. Compared with the decision-making model from a single view, multi-view learning is capable of exploiting abundant information from various views to enhance decision-making quality and model robustness. Multi-view learning has seen tremendous works in both supervised \cite{han2022trusted,xu2024reliable,shi2024generalized,zhang2024discovering} and unsupervised paradigms \cite{liu2018late,lin2022dual,lin2021multi}. 

Recently, information theory has been shown to offer powerful representations for multi-view data, with particular attention on Information Bottleneck (IB) \cite{alemi2016deep,yu2021information,hu2024survey}. IB aims to extract the most compressed representation of the input while retaining task-relevant information. 
As one of the earliest attempts to incorporate the Information Bottleneck (IB) principle into multi-view representation learning, Multi-view Information Bottleneck (MIB) \cite{federici2020learning} learns robust, unsupervised representations by maximizing the mutual information between encoded views while minimizing the entropy of view-specific information. In contrast to MIB, which is grounded specifically in the IB framework, subsequent methods have adopted a more diverse information-theoretic perspective to guide representation learning \cite{wan2021multi,hu2020dmib,wang2023self,cui2024novel}.
Through a profound understanding of these studies, we identify the following observations and corresponding shortcomings. 1) These methods rely on an important assumption called {\it multi-view redundancy}: the common information between views is necessary and sufficient for down-stream tasks \cite{federici2020learning}. With this assumption, only the common information is exploited for prediction; however, this assumption overlooks the potential predictive capabilities of view-unique information for tasks. 2) Many methods exhibit noteworthy limitations when dealing with more than two views \cite{wang2023self}. They are either designed specifically for two views, or are based on relying on viewpair common information, rather than the truly representative common information across multiple views.

To address the aforementioned drawbacks, we propose a Comprehensive Information-theoretic Multi-view Learning framework (CIML). 
CIML is mainly composed of two modules: a consistency-learning module and a uniqueness-learning module. The consistency-learning module aims to capture task-relevant information shared among all views. Specifically, it first maximizes the G$\acute{a}$cs-K$\mathaccent"707F o$rner common information to extract latent representations that can be jointly reconstructed from each view; then, an IB objective function is enforced to compress this common information while retaining only the most predictive features for the downstream task. The uniqueness-learning module focuses on discovering unique and predictive information in each individual view. For each view, it applies the IB principle to learn the most compressed unique representation to preserve the task-relevant signal. Meanwhile, two explicit constraints are imposed: (1) minimizing the mutual information between each unique representation and the common representation to ensure disjointness; (2) minimizing the mutual information between any pair of unique representations from different views to ensure pure uniqueness. Consequently, CIML contributes to a more comprehensive framework by explicitly discarding the {\it multi-view redundancy} assumption, yielding a predictively sufficient representation for downstream tasks.
Our main contributions include:

\begin{itemize}
\item {Common representation learning}: We first extract shared features by maximizing G$\acute{a}$cs-K$\mathaccent"707F o$rner common information. Then, we compress the learned common information to achieve a sufficient yet task-relevant representation with IB. 

\item {Unique representation learning}: We employ IB again to learn the most compressed unique representation for each view, while simultaneously imposing two constraints: minimizing the mutual information between unique and common representations, as well as among different unique representations. 

\item {We theoretically prove that the joint representation is predictively sufficient, demonstrating its comprehensiveness in capturing information relevant to downstream task prediction. Experimental results on real-world datasets demonstrate that CIML outperforms several state-of-art models.} 
\end{itemize}  

The remainder of this paper is organized as follows. Section~\RN{2} presents related works. Section~\RN{3} shows the detailed procedures of CIML. Section~\RN{4} carries out extensive experiments to validate CIML's performance. We finally draw some conclusions in Section~\RN{5}.

\section{Related Work}

\subsection{Multi-view Learning}
Multi-view learning aims to effectively leverage data from multiple views to handle unsupervised, supervised, and semi-supervised tasks. Many early studies focused on learning the self-representation matrix in a low-dimensional subspace \cite{gao2015multi,cao2023robust}. However, most of them directly averaged different learned self-representation matrices to achieve a common self-representation matrix, which probably fail to make use of complementary information among different views. To better explore the comprehensive information of multi-view data, the latent representation-based approach that projects multiple views into a shared latent space has been extensively investigated \cite{zhang2018generalized,chen2020multi,shi2024enhanced}. In addition, tensor-based multi-view learning methods, which store the representation matrix of each view as the frontal slice of a third-order tensor, are capable of exploiting the high-order correlations among views \cite{xie2018unifying,wen2021unified,chang2024tensorized,qin2023flexible}. Recent research has increasingly focused on low-rank tensor techniques \cite{guo2022logarithmic,wang2025tensorized}, as they offer a promising approach for integrating both spatial and comprehensive information (i.e., consensus and complementary information) within a unified framework, while simultaneously mitigating redundant and noisy data. 

The key of multi-view learning lies in how to exploit data information, whether from an intra-view or inter-view perspective. As an alternative approach to achieve this goal, various efforts have also been devoted to graph-based \cite{wang2019gmc,tan2023sample,pan2021multi} and deep learning methods \cite{wang2015deep,lin2021completer}. Generally, graph-based multi-view methods focus on the quality of the constructed graph, while deep learning methods emphasize exploring the underlying non-linear structure of multi-view data. Recently, there has been growing recognition within the research community of the importance of addressing incomplete multi-view learning scenarios \cite{qin2022nim, chen2024spectral, yang2025tensor}, particularly in terms of exploring data completion and reconstruction techniques \cite{fu2024anchor,liu2025reliable}. These approaches enable the learning of complex representations from incomplete data, making them effective in situations characterized by high levels of missing or noisy information.

\subsection{Information-theoretic Learning}
As a widely popular information theory method in recent years, the IB principle provides a paradigm that compresses data by retaining only the most relevant information for a given task, thereby enhancing robustness and generalization \cite{kawaguchi2023does}. Upon utilizing variational inference to construct a lower bound for computing mutual information, Variational IB (VIB) makes it possible to train IB-guided deep neural networks. In the case of two views, Federici \emph{et al}. proposed MIB for robust representation learning by identify superfluous information as that not shared by both views \cite{federici2020learning}. Inspired by MIB, subsequent studies have made significant efforts towards further exploring the predictive capabilities from original features or data representations \cite{wan2021multi,cui2024novel,huang2023generalized,wang2023self}. While these methods depend on variational approximation for mutual information estimation, Yan \emph{et al.} proposed a differentiable approach that directly fits mutual information, thereby addressing the challenge of estimation in high-dimensional multi-view spaces \cite{yan2024differentiable}. From an information-theoretic optimization standpoint, further progress has been made through a framework that maximizes intra-view coding rate reduction together with inter-view mutual information \cite{yan2025multiview}. In addition, Lou \emph{et al.} proposed a self-supervised weighted IB method to more effectively capture the complementary information across views \cite{lou2024self}.

However, almost the above mentioned methods rely on an important assumption called \emph{multi-view redundancy}: the common information between views is necessary and sufficient for down-steam tasks (see \cite{lin2022dual} for typical examples, where the Dual Contrastive Prediction (DCP) framework is introduced); however, this assumption naturally ignores the potential predictive capabilities of view-unique information. Additionally, most existing methods are specifically designed for the two-view case or focus on learning shared information for view pairs, rather than truly common information across multiple views (see \cite{wang2023self} for typical examples). Beyond the IB principle, G$\acute{a}$cs-K$\mathaccent"707F o$rner (GK) common information—also known as ``zero error information" \cite{wolf2004zero}—offers an alternative perspective by quantifying the shared information between two random variables through the largest reconstructable common part \cite{gacs1973common}. This concept has been recently applied to modify the traditional Variational Auto-Encoder (VAE), making it more suitable for high-dimensional samples \cite{kleinman2023gacs}.  

\begin{figure*}[htbp]
	\centering
	\includegraphics[scale=0.535]{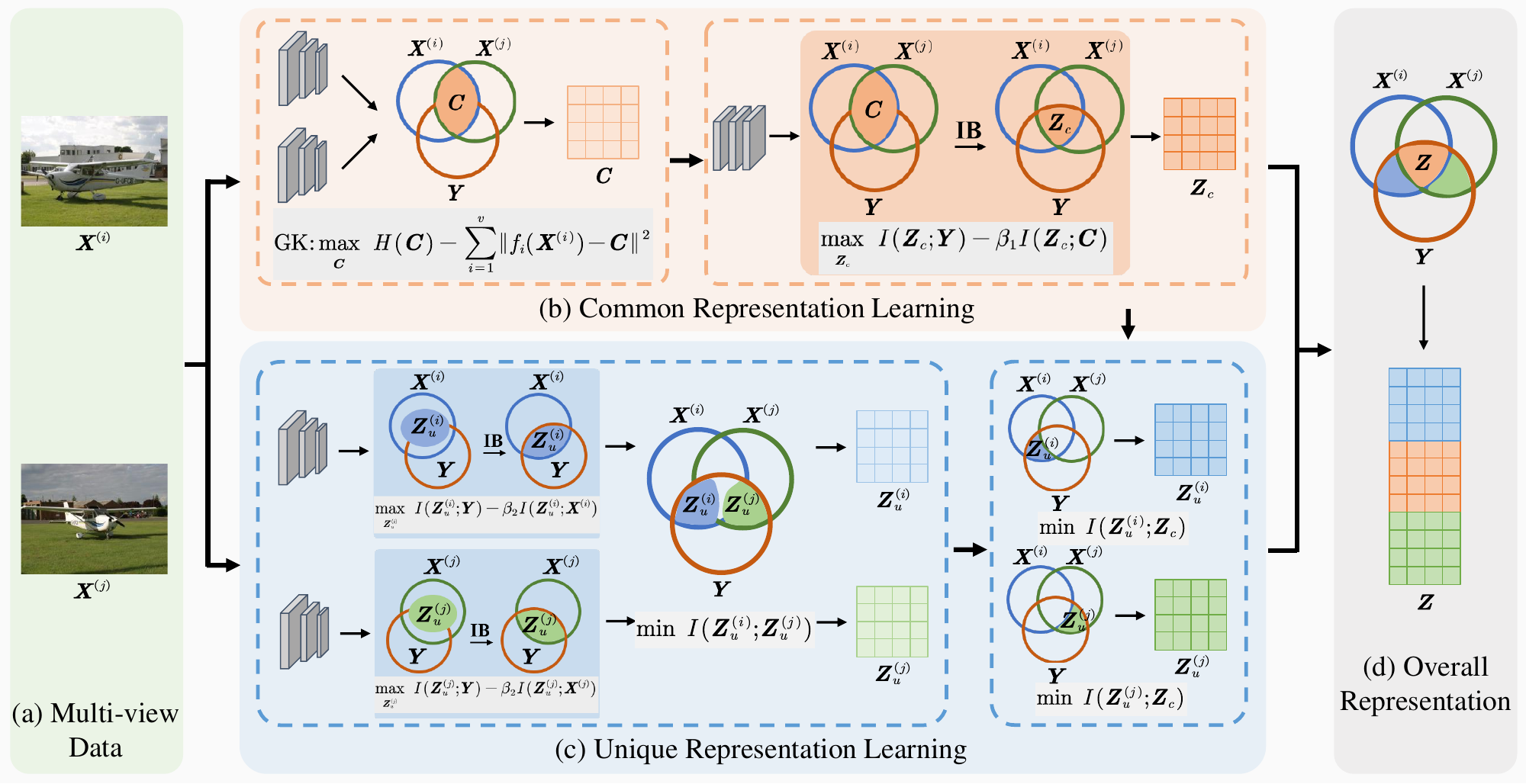} 
	\caption{Framework of CIML.}
	\label{framework}
\end{figure*} 

\section{Proposed Methodology}

In this section, we present the learning mechanisms of CIML, including both common and unique representations. Consider multi-view data $\{\boldsymbol{X}^{(i)}\}_{i=1}^v$ with $n$ samples and $m$ classes, where $\boldsymbol{X}^{(i)} = [\boldsymbol{x}^{(i)}_1, \boldsymbol{x}^{(i)}_2,\cdots, \boldsymbol{x}^{(i)}_n]\in \mathbb{R}^{d_i\times n}$ represents the $i$-th view data with $d_i$ dimension, with $\boldsymbol{x}^{(i)}_k$ denoting the $k$-th instance of the $i$-th view. Fig. \ref{framework} shows an illustration for two views, where $\boldsymbol{Y}$ corresponds to a downstream task. For learning common representation, we first employ G$\acute{a}$cs-K$\mathaccent"707F o$rner common information to identify shared features. Based on IB, we then eliminate superfluous information that does not contribute to the downstream task, retaining only the sufficient task-relevant representations. For learning unique representations, we hope that each view maintains sufficient distinguishing features that set it apart from other views, while ensuring no overlap with the common representation to guarantee distinctiveness. Finally, we obtain a complete representation by integrating both common and view-unique representations. In the following, we present the motivation behind our research, provide a detailed description of each module, as well as an insightful discussion.

\subsection{Research Motivation}

To the best of our knowledge, there is little prior work that explicitly discards the assumption of \emph{multi-view redundancy}. Most IB-based methods for multi-view learning—such as MIB and DCP—adhere to this assumption. While a few works relax this assumption, they address isolated aspects—using IB mainly to encourage compact representations (e.g., CMIB-Nets \cite{wan2021multi}) or extracting only pairwise common information—rather than providing a comprehensive information-theoretic framework. In contrast, our model is the first to discard the \emph{multi-view redundancy} assumption while constructing a complete information-theoretic framework (going beyond IB) for multi-view learning by jointly leveraging the predictive capabilities of both common and specific information.

\subsection{Common Representation Learning}

In determining the learning mechanism of common representation, our primary consideration is that it should effectively capture the truly common information when the number of views exceeds two, rather than merely averaging the common information from view-pairs as the final representation. To reach this target, the essence lies in finding the underlying latent representation that ensures the encoded features of each view are as close to it as possible. This demand naturally corresponds to G$\acute{a}$cs-K$\mathaccent"707F o$rner common information, which aims to identify the maximum amount of information that can be jointly observed by multiple sources from their respective observations. Formally, the common information $\boldsymbol{C}$ among multiple views can be learned via $GK(\boldsymbol{X}^{(1)},\cdots, \boldsymbol{X}^{(v)})$:
\begin{equation}
	\underset{\boldsymbol{C}}{max} \;\; H(\boldsymbol{C}) - \sum^{v}_{i=1} {\lVert f_i(\boldsymbol{X}^{(i)}) - \boldsymbol{C} \rVert^{2}},
	\label{eq01}
\end{equation}
where $H(\cdot)$ denotes the information entropy, $\lVert\cdot\rVert$ takes the square of $l_2$-norm, and $f_i(\cdot)$ serves to encode the original data from the $i$-th view. Eq. (\ref{eq01}) achieves the maximum amount of information (i.e., $H(\boldsymbol{C})$), subject to minimizing the discrepancy between the underlying common information and each encoded input feature. 


After obtaining the common information $\boldsymbol{C}$, we seek to obtain a more compressed representation that is sufficient for predicting downstream tasks. This inherently motivates us to employ IB for further processing, resulting in:
\begin{equation}
	\underset{\boldsymbol{Z}_c}{max} \;\; I(\boldsymbol{Z}_c; \boldsymbol{Y}) - \beta_1 
I(\boldsymbol{Z}_c; \boldsymbol{C}),
	\label{eq02}
\end{equation}
where $I(A;B)$ describes the mutual information between variables $A$ and $B$, and $\beta_1$ is a trade-off parameter. With Eq. (\ref{eq02}), we can retain only the most relevant common representation for $\boldsymbol{Y}$ while discarding redundant information. The optimization involves two terms related to the calculation of mutual information: $I(\boldsymbol{Z}_c; \boldsymbol{Y})$ and $I(\boldsymbol{Z}_c; \boldsymbol{C})$. According to the definition of mutual information \cite{bachman2019learning}, the first term $I(\boldsymbol{Z}_c; \boldsymbol{Y})$ can be expressed as
\begin{equation}
	\begin{aligned}
I(\boldsymbol{Z}_c; \boldsymbol{Y}) &= \iint p(\boldsymbol{z}_c,\boldsymbol{y}) \log \frac{p(\boldsymbol{z}_c, \boldsymbol{y})}{p(\boldsymbol{z}_c)p(\boldsymbol{y})} d\boldsymbol{y} d\boldsymbol{z}_c \\
         &= \iint p(\boldsymbol{z}_c,\boldsymbol{y}) \log \frac{p(\boldsymbol{y}|\boldsymbol{z}_c)}{p(\boldsymbol{y})} d\boldsymbol{y} d\boldsymbol{z}_c. \\
	\end{aligned}
	\label{eq03}
\end{equation}  

It is intractable to directly compute the above equation. As an alternative approach, recent studies attempted to optimize the variational lower bounds of the original objective function to derive an approximate solution {\cite{fabius2014variational}. Specifically, let $q(\boldsymbol{y}|\boldsymbol{z}_c)$ be a variational approximation to $p(\boldsymbol{y}|\boldsymbol{z}_c)$, and introduce the definition of Kullback-Leibler (KL) divergence between $p(\boldsymbol{y}|\boldsymbol{z}_c)$ and $q(\boldsymbol{y}|\boldsymbol{z}_c)$
\begin{equation}
\begin{aligned}
D_{KL}(p(\boldsymbol{y}|\boldsymbol{z}_c) \parallel q(\boldsymbol{y}|\boldsymbol{z}_c)) &= \iint p(\boldsymbol{y}|\boldsymbol{z}_c) \log \frac{p(\boldsymbol{y}|\boldsymbol{z}_c)}{q(\boldsymbol{y}|\boldsymbol{z}_c)} d\boldsymbol{y}\,d\boldsymbol{z}_c. 
\end{aligned}
\label{eq04}
\end{equation}
Since $D_{KL}(p(\boldsymbol{y}|\boldsymbol{z}_c) \parallel q(\boldsymbol{y}|\boldsymbol{z}_c)) \geq 0$, from Eq. (\ref{eq04}) we easily have
\begin{equation}
\begin{aligned}
	&\iint p(\boldsymbol{z}_c, \boldsymbol{y}) \log p(\boldsymbol{y}|\boldsymbol{z}_c) d\boldsymbol{y} d\boldsymbol{z}_c \geq \\
	&\iint p(\boldsymbol{z}_c, \boldsymbol{y}) \log q(\boldsymbol{y}|\boldsymbol{z}_c) d\boldsymbol{y} d\boldsymbol{z}_c.
\end{aligned}
\label{eq05}
\end{equation}

\noindent By combing Eqs. (\ref{eq03}) and (\ref{eq05}), we derive 
\begin{equation}
\begin{aligned}
I(\boldsymbol{Z}_c; \boldsymbol{Y}) & \geq \iint p(\boldsymbol{z}_c,\boldsymbol{y}) \log \frac{q(\boldsymbol{y}|\boldsymbol{z}_c)}{p(\boldsymbol{y})} d\boldsymbol{y}d\boldsymbol{z}_c \\
         &= \iint p(\boldsymbol{z}_c,\boldsymbol{y}) \log q(\boldsymbol{y}|\boldsymbol{z}_c) d\boldsymbol{y} d\boldsymbol{z}_c + H(\boldsymbol{y}) \\
         & \geq \iint p(\boldsymbol{z}_c,\boldsymbol{y}) \log q(\boldsymbol{y}|\boldsymbol{z}_c) d\boldsymbol{y} d\boldsymbol{z}_c. \\
\end{aligned}
\label{eq06}
\end{equation}

\noindent Due to $p(\boldsymbol{z}_c,\boldsymbol{y}) = \int p(\boldsymbol{x})p(\boldsymbol{y}|\boldsymbol{x})p(\boldsymbol{z}_c|\boldsymbol{x}) d\boldsymbol{x}$, the above inequality can be further formulated as
\begin{equation}
\begin{aligned}
	I(\boldsymbol{Z}_c;\boldsymbol{Y}) &\geq \iiint p(\boldsymbol{x})p(\boldsymbol{y}|\boldsymbol{x})p(\boldsymbol{z}_c|\boldsymbol{x}) \log q(\boldsymbol{y}|\boldsymbol{z}_c) d\boldsymbol{x} d\boldsymbol{y} d\boldsymbol{z}_c\\
	&= \iint p(\boldsymbol{x})p(\boldsymbol{y}|\boldsymbol{x})d\boldsymbol{x}d\boldsymbol{y} \int p(\boldsymbol{z}_c|\boldsymbol{x}) \log q(\boldsymbol{y}|\boldsymbol{z}_c) d\boldsymbol{z}_c.
\end{aligned}
	\label{eq07}
\end{equation}

\noindent In practice, we can approximate the double integration over $\boldsymbol{x}$ and $\boldsymbol{y}$ by Monte Carlo sampling \cite{alemi2016deep}, allowing us to obtain an empirically computable lower bound:
\begin{equation}
	I(\boldsymbol{Z}_c;\boldsymbol{Y}) \approx \frac{1}{N} \sum^N_{k=1} \int p(\boldsymbol{z}_c|\boldsymbol{x}_k) \log q(\boldsymbol{y}_k|\boldsymbol{z}_c) \, d\boldsymbol{z}_c,
\label{eq08}
\end{equation}
where $N$ denotes the size of sampled data.

Similarly, let $r(\boldsymbol{z}_c)$ be a variational approximation to $p(\boldsymbol{z}_c)$; thus, we obtain a variational bound for the second term $I(\boldsymbol{Z}_c; \boldsymbol{C})$ 
\begin{equation}
\begin{aligned}
I(\boldsymbol{Z}_c;\boldsymbol{C}) &\approx \frac{1}{N} \sum^N_{k=1}{\int p(\boldsymbol{z}_c|\boldsymbol{c}_k) \log \frac{p(\boldsymbol{z}_c|\boldsymbol{c}_k)}{r(\boldsymbol{z}_c)}} \, d\boldsymbol{z}_c \\
         &= \frac{1}{N} \sum^N_{j=k}{D_{KL}(p(\boldsymbol{z}_c|\boldsymbol{c}_k) \parallel r(\boldsymbol{z}_c))}. \\
\end{aligned}
\label{eq09}
\end{equation}
Readers are encouraged to refer to Appendix A for detailed procedures on deriving Eq. (\ref{eq09}). Based on the aforementioned discussion, the loss function for learning common representation is given by
\begin{equation}
\begin{aligned}
	\mathcal{L}_c = &-H(\boldsymbol{C}) + \sum^{v}_{i=1} {\lVert f_i(\boldsymbol{X}^{(i)}) - \boldsymbol{C} \rVert^{2}} \\
	&- [I(\boldsymbol{Z}_c;\boldsymbol{Y}) - \beta_1 I(\boldsymbol{Z}_c;\boldsymbol{C})].
\end{aligned}
	\label{eq10}
\end{equation}

\subsection{Unique Representation Learning}

In learning the unique representation for each view, we aim to contain only task-relevant information. Meanwhile, to ensure pure uniqueness, the following criteria should be met: 1) the unique representations between any two views should be distinct, and 2) each unique representation should not overlap with the learned common representation. 

To achieve this goal, we formulate the following optimization problem
\begin{equation}
\begin{aligned}
	&\underset{\boldsymbol{Z}^{(i)}_u}{max} \;\; \sum^v_{i=1} I(\boldsymbol{Z}^{(i)}_u; \boldsymbol{Y}) - \beta_2 I(\boldsymbol{Z}^{(i)}_u; \boldsymbol{X}^{(i)}),\\   
\end{aligned}
	\label{eq11}
\end{equation}
subject to
\begin{equation}
	min \;\; \sum_{i=1}^v \sum_{j=1,j\neq i}^{v}{I(\boldsymbol{Z}^{(i)}_u;\boldsymbol{Z}^{(j)}_u)} \;\; \mathrm{and} \;\; min \;\; \sum_{i=1}^{v} I(\boldsymbol{Z}^{(i)}_u; \boldsymbol{Z}_c).
	\label{eq12}
\end{equation}

It is seen that solving the above optimization problem involves the calculation of four mutual information terms, namely $I(\boldsymbol{Z}^{(i)}_u; \boldsymbol{Y})$, $I(\boldsymbol{Z}^{(i)}_u; \boldsymbol{X}^{(i)})$, $I(\boldsymbol{Z}^{(i)}_u;\boldsymbol{Z}^{(j)}_u)$, and $I(\boldsymbol{Z}^{(i)}_u; \boldsymbol{Z}_c)$. Specifically, $I(\boldsymbol{Z}^{(i)}_u; \boldsymbol{Y})$ and $I(\boldsymbol{Z}^{(i)}_u; \boldsymbol{X}^{(i)})$ can be computed following the same procedures as those used for calculating $I(\boldsymbol{Z}_c;\boldsymbol{Y})$ and $I(\boldsymbol{Z}_c;\boldsymbol{C})$. Let $q(\boldsymbol{y}|\boldsymbol{z}^{(i)}_u)$ and $r(\boldsymbol{z}^{(i)}_u)$ be variational approximations to $p(\boldsymbol{y}|\boldsymbol{z}^{(i)}_u)$ and $p(\boldsymbol{z}^{(i)}_u)$, respectively, we easily arrive at the corresponding variational bounds for the $i$-th view
\begin{equation}
	I(\boldsymbol{Z}_u^{(i)};\boldsymbol{Y}) \approx \frac{1}{N} \sum^N_{k=1} \int p(\boldsymbol{z}_u^{(i)}|\boldsymbol{x}_k) \log q(\boldsymbol{y}_k|\boldsymbol{z}_u^{(i)}) \, d\boldsymbol{z}_u^{(i)},
	\label{eq13}
\end{equation}
and
\begin{equation}
 	I(\boldsymbol{Z}^{(i)}_u;\boldsymbol{X}^{(i)}) \approx \frac{1}{N} \sum^N_{k=1}{D_{KL}(p(\boldsymbol{z}_u^{(i)}|\boldsymbol{x}_k^{(i)}) \parallel r(\boldsymbol{z}_u^{(i)}))}.
 	\label{eq14}
\end{equation}

In addition, for $I(\boldsymbol{Z}^{(i)}_u;\boldsymbol{Z}^{(j)}_u)$ and $I(\boldsymbol{Z}^{(i)}_u; \boldsymbol{Z}_c)$, we can use the MINE estimator \cite{belghazi2018mine} that performs sampling from the respective variational encoders of the corresponding variables to find their variational bounds:
\begin{equation}
	I(\boldsymbol{Z}_u^{(i)};\boldsymbol{Z}^{j}_u) \geq I_{MINE}(\boldsymbol{Z}_u^{(i)};\boldsymbol{Z}^{j}_u),
	\label{eq15}
\end{equation}  
and 
\begin{equation}
	I(\boldsymbol{Z}_u^{(i)};\boldsymbol{Z}_c) \geq I_{MINE}(\boldsymbol{Z}_u^{(i)};\boldsymbol{Z}_c).
	\label{eq16}
\end{equation}

From Eqs. (\ref{eq11}) to (\ref{eq12}), we have the following loss function for learning unique representation
\begin{equation}
\begin{aligned}
	\mathcal{L}_u = -&\sum^{v}_{i=1} {[I(\boldsymbol{Z}^{(i)}_u;\boldsymbol{Y}) - \beta_2 I(\boldsymbol{Z}^{(i)}_u;\boldsymbol{X}^{(i)}) - I(\boldsymbol{Z}^{(i)}_u;\boldsymbol{Z}_c)]} \\
	&+ \sum_{i=1}^v \sum_{j=1,j\neq i}^{v}{I(\boldsymbol{Z}^{(i)}_u;\boldsymbol{Z}^{(j)}_u)}.
\end{aligned}
	\label{eq17}
\end{equation} 

\subsection{Model Loss}

For classification tasks, we need to define the cross-entropy loss $\mathcal{L}_{ce}$ 
\begin{equation}	
	\mathcal{L}_{ce} = -\sum_{k=1}^n \sum_{j=1}^m {y}_{kj}
    \log(\hat{{y}}_{kj}). 
	\label{eq18}
\end{equation}
 
By combining Eq. (\ref{eq18}) and the loss functions for learning both common and unique representations, the total loss function of our model is expressed as:

\begin{equation}
	\mathcal{L} = \mathcal{L}_{ce} + \beta_3 \mathcal{L}_c + \beta_4 \mathcal{L}_u,    
	\label{eq19}	
\end{equation}
where $\beta_3$ and $\beta_4$ are two hyperparameters. Although our model involves four hyperparameters (including the additional $\beta_1$ and $\beta_2$ for learning common and unique representations), it is important to note that $\beta_1$ and $\beta_2$ are derived from the IB principle, where they regulate the degree of compression in the learned representations. In practice, these parameters are typically assigned small values. Our parameter sensitivity analysis in the experimental section further confirms this choice. Therefore, unless otherwise specified, we fix both $\beta_1$ and $\beta_2$ to $10^{-4}$ throughout this paper. The implementation procedures of our model is presented in Algorithm \ref{alg:CIML}.

\begin{algorithm}[tb]
    \caption{Implementation Procedures of CIML}
    \label{alg:CIML}
    \textbf{Input}: Multi-view data \{$\boldsymbol{X}^{(1)},\boldsymbol{X}^{(2)},...,\boldsymbol{X}^{(v)}$\}, corresponding labels $\boldsymbol{Y}$, hyperparameters $\beta_1$, $\beta_2$, $\beta_3$ and $\beta_4$.\\
    \textbf{Initialize}: Autoencoder parameters $\{\boldsymbol{\Theta}_{c}^{(i)}, \boldsymbol{\Theta}_{u}^{(i)}\}_{i=1}^{v}$, $\boldsymbol{\Theta}_{\boldsymbol{C}}$ and $\boldsymbol{C}$\\
    \textbf{Output}: Comprehensive representation $\boldsymbol{Z}$ and predicted labels $\widehat{\boldsymbol{Y}}$.
    \begin{algorithmic}[1] 
        \WHILE{not converged}
            \STATE Obtain $\boldsymbol{Z}_c$ by encoding $\boldsymbol{C}$ with common representation Autoencoder;
            \FOR{$i = 1$ \TO $v$}
                \STATE Obtain $f_i(\boldsymbol{X}^{(i)})$ with common representation Autoencoder and $\boldsymbol{Z}_u^{(i)}$ with unique representation Autoencoder;
            \ENDFOR
            \STATE Calculate common representation loss $\mathcal{L}_c$ by Eq. (\ref{eq10});
            \STATE Calculate unique representation loss $\mathcal{L}_u$ by Eq. (\ref{eq18});
            \STATE Combine $\boldsymbol{Z}_u^{(i)}$ with $\boldsymbol{Z}_c$ to form final feature vector $\boldsymbol{Z}$;
            \STATE Predict output $\widehat{\boldsymbol{Y}}$ using classifier with representation $\boldsymbol{Z}$;
            \STATE Calculate total loss $\mathcal{L}$ by Eq. (\ref{eq19});
            \STATE Update $\{\boldsymbol{\Theta}_{c}^{(i)}, \boldsymbol{\Theta}_{u}^{(i)}\}_{i=1}^{v}$, $\boldsymbol{\Theta}_{\boldsymbol{C}}$ and $\boldsymbol{C}$ by total loss $\mathcal{L}$;
        \ENDWHILE
        \RETURN Comprehensive representation $\boldsymbol{Z}$ and predicted labels $\widehat{\boldsymbol{Y}}$.
    \end{algorithmic}
\end{algorithm}

\subsection{Computational Complexity Analysis}
Let $N$ denote the number of samples in the dataset, $B$ the batch size, $D_v = \sum_{i=1}^{V} d_i$ the total dimensionality across all views, $D_z$ the dimensionality of the learned representation $Z$, and $D_h$ the dimensionality of the latent representation. For $iter$ iterations, the computational cost for the consistent component is $O(iter N D_v D_z)$ and for the unique component is $O(iterN V^2 D_z^2)$. The overall cost can be approximated as $O(iter N D_v D_z + iter N V^2 D_z^2)$, as showing in Table \ref{Time Complexity}, which is comparable to the computational complexity of existing IB-based methods. 


\begin{table}[!htb]
    \centering
    \small
    \caption{Comparison of Computational Complexity.}
    \begin{tabular}{lc}
    \hline
        Method & Computational Complexity \\
    \hline
    	MIB    & $O(iter N D_v D_z)$ \\
    	CMIB-Nets & $O(iter N V D_h(D_v + D_z)$ \\
    	DCP    & $O(iter N V D_z(D_v + V + B)$ \\
        F\scalebox{0.85}{ACTOR}CL & $O(iter N D_z(B + D_v + D_z)$ \\
        CUMI   & $O(iter N^2 V + N V D_v D_z)$ \\
        Ours   & $O(iter N V D_z (D_v + V D_z))$ \\
    \hline
    \end{tabular}
    \label{Time Complexity}
\end{table}

\subsection{Discussion}

\subsubsection{Comprehensiveness of Representation}

We argue that the learned joint representation $\boldsymbol{Z} = (\boldsymbol{Z}_c, \boldsymbol{Z}^{(1)}_u, \dots, \boldsymbol{Z}^{(v)}_u)$ demonstrates comprehensiveness with respect to the prediction task, in the sense that it retains all information from the original views relevant to predicting the target $\boldsymbol{Y}$. Formally, we define this property as \emph{predictive sufficiency}.\\
\textbf{Theorem 3.1 (Predictive Sufficiency)} The joint representation $\boldsymbol{Z} = (\boldsymbol{Z}_c, \boldsymbol{Z}^{(1)}_u, \dots, \boldsymbol{Z}^{(v)}_u)$ is predictively sufficient for the target $\boldsymbol{Y}$ if, given that the original views $\boldsymbol{X}=(\boldsymbol{X}^{(1)}, \cdots, \boldsymbol{X}^{(v)})$ fully determine $\boldsymbol{Y}$, the joint representation $\boldsymbol{Z}$ approximately determines $\boldsymbol{Y}$. 

This implies that the uncertainty in $\boldsymbol{Y}$ given $\boldsymbol{Z}$ is nearly equal to that given the full input $\boldsymbol{X}$, indicating that $\boldsymbol{Z}$ retains most of the predictive information contained in $\boldsymbol{X}$.


\emph{Proof}.
Please see Appendix B for proof.

\subsubsection{Comparison with Existing Methods}

Prior work has extensively explored information-theoretic approaches to multi-view representation learning. While these methods have made significant process, we argue that our framework offers a more comprehensive formulation by relaxing the commonly used assumption of \emph{multi-view redundancy}. Unlike existing approaches that rely on this constraint, such as \cite{federici2020learning,lin2022dual}, our framework explicitly accouts for the predictive power of view-specific representations, leading to a more flexible and task-aware learning process. As a result, our model can be naturally regarded as advancing beyond these methods. In light of this, we proceed to highlight the key distinctions between our approach and recent methods that also aim to discard the \emph{multi-view redundancy} assumption.

To the best of our knowledge, there are two typical methods that explore the relaxation of the \emph{multi-view redundancy assumption}: \textbf{IMC} \cite{huang2023generalized} and \textbf{F\scalebox{0.85}{ACTOR}CL} \cite{liang2023factorized}. Notably, both IMC and F\scalebox{0.85}{ACTOR}CL are limited in their ability to extract shared information when more than two views are available, as they fundamentally focus on capturing pairwise common information between views. In contrast, our CIML framework captures GK common information, which more accurately characterizes the underlying shared features across all views. This constitutes a key distinction between our model and IMC or F\scalebox{0.85}{ACTOR}CL. Below, we further elaborate on the specific differences between our framework and each of these two approaches.

\textbf{Comparison with IMC} \; IMC aims to maximize the mutual information between a shared representation and input views, while minmizing view-specific mutual information. IMC is an implicit approach that relies on conditional mutual information terms, such as $\min I(\boldsymbol{Z}^{(1)}; \boldsymbol{Z}^{(2)} | \boldsymbol{Z})$, to separate shared and unique information. However, estimating such terms often demands high computational costs, limiting the scalability of the method. In contrast, our model explicitly determine the learning framework for the common representation $\boldsymbol{Z_c}$ and view-specific representations $\boldsymbol{Z}^{(i)}_u$.

\textbf{Comparison with F\scalebox{0.85}{ACTOR}CL} \; Despite that both F\scalebox{0.85}{ACTOR}CL and our model explicitly defines shared and uniqe representation for learning, they differ significantly in two aspects. First, F\scalebox{0.85}{ACTOR}CL is a contrastive learning-based framework, whereas our model is a fully information-theoretic framework. Second, F\scalebox{0.85}{ACTOR}CL} directly factorizes task-relevant information into shared and unique components from the original view inputs, while our CIML model employs a compressed process to learn these representations. Importantly, our model enforces an independence constraint between unique representations from different views to ensure their purity, i.e., $min \;\; \sum_{i=1}^v \sum_{j=1,j\neq i}^{v}{I(\boldsymbol{Z}^{(i)}_u;\boldsymbol{Z}^{(j)}_u)}$, which F\scalebox{0.85}{ACTOR}CL does not enforce.      

\section{Experiments}

\subsection{Experimental Settings}

\subsubsection{Datasets}

We select six real-world multi-view datasets. \textbf{MSRC-v1}\footnote{https://mldta.com/dataset/msrc-v1/} contains 30 images per class, encompassing a total of 7 different classes. Six types of features are extracted. \textbf{LandUse-21}\footnote{http://weegee.vision.ucmerced.edu/datasets/landuse.html} contains 2,100 satellite images from 21 categories, and we use the GIST, PHOG and LBP features as three views. \textbf{Caltech101-20}\footnote{https://data.caltech.edu/records/mzrjq-6wc02} is a subset of the renowned Caltech101 image classification dataset, which consists of 2,386 images of 20 subjects and includes six features. \textbf{NUS} \cite{chua2009nus} consists of 81 concepts, 269,648 images about animal concept. We select 12 categories and totally 2400 data samples which compose of cat, cow, dog, elk, hawk, horse, lion, squirrel, tiger, whales, wolf and zebra. \textbf{Scene15}\footnote{https://www.kaggle.com/zaiyankhan/15scene-dataset} contains 4,485 images of 15 indoor and outdoor scene categories. GIST, PHOG, and LBP features are extracted as three views. \textbf{NoisyMNIST} \cite{wang2015deep} contains 50,000 training images. It is generated by introducing white Gaussian noise, motion blur and a combination of additive white Gaussian noise and reduced contrast to the MNIST dataset. We randomly select 20,000 samples.

\begin{table}[!htb]
    \centering
    \small
    \caption{Summary of multi-view datasets}    
    \begin{tabular}{lrrrr}
    \hline
        Dataset & View & Sample & Dimension of features \\
    \hline
    	MSRC-v1       & 5 & 210   & {24, 576, 512, 256, 254} \\
    	LandUse-21    & 3 & 2100  & {20, 59, 40} \\
        Caltech101-20 & 6 & 2386  & {48, 40, 254, 1984, 512, 928} \\
        NUS           & 6 & 2400  & {64, 144, 73, 128, 225, 500} \\
        Scene-15      & 3 & 4485  & {20, 59, 40} \\
        NoisyMINST    & 2 &	20000 & {784, 784} \\
    \hline
    \end{tabular}
    \label{tab:dataset}
\end{table}

\begin{table*}
    \centering
    \caption{Classification results on the MSRC-v1, LandUse-21, and Caltech101-20 datasets. Bold font denotes the best performance, and underline font indicates the second best performance.}
    \resizebox{\linewidth}{!}{
    \begin{tabular}{lccccccccc}
    \hline
        Datasets & \multicolumn{3}{c}{MSRC-v1} & \multicolumn{3}{c}{LandUse-21} & \multicolumn{3}{c}{Caltech101-20} \\
    \cmidrule(lr){2-4} \cmidrule(lr){5-7} \cmidrule(lr){8-10}
        Method & ACC (\%) & Precision (\%) & F1 (\%) & ACC (\%) & Precision (\%) & F1 (\%) & ACC (\%) & Precision (\%) & F1 (\%) \\
    \hline
        DCCA           & 92.14 $\pm$ 3.20 & 93.34 $\pm$ 3.14 & 92.32 $\pm$ 3.11 & 60.50 $\pm$ 1.74 & 61.95 $\pm$ 2.14 & 60.31 $\pm$ 1.74 & 84.60 $\pm$ 1.15 & 85.60 $\pm$ 1.20 & 84.52 $\pm$ 1.13 \\
        MIB             & 94.76 $\pm$ 1.43 & 95.19 $\pm$ 1.33 & 94.69 $\pm$ 1.50 & 43.00 $\pm$ 0.92 & 44.34 $\pm$ 2.33 & 41.39 $\pm$ 1.06 & 87.05 $\pm$ 1.00 & 87.01 $\pm$ 1.21 & 85.66 $\pm$ 1.19  \\ 
        TMC             & 95.48 $\pm$ 0.71 & \underline{97.10 $\pm$ 0.30} & \underline{97.10 $\pm$ 0.30} & 61.00 $\pm$ 4.90 & 67.10 $\pm$ 4.35 & 41.40 $\pm$ 6.48 & 92.82 $\pm$ 1.03 & \textbf{97.50 $\pm$ 0.67} & 85.70 $\pm$ 1.27  \\
        DUA-Nets        & 81.43 $\pm$ 3.96 & 81.69 $\pm$ 5.58 & 80.11 $\pm$ 4.56 & 57.74 $\pm$ 2.17 & 63.76 $\pm$ 2.53 & 56.23 $\pm$ 2.46  & 84.34 $\pm$ 1.18 & 84.90 $\pm$ 2.36 & 69.14 $\pm$ 2.73\\
        CMIB-Nets       & 80.00 $\pm$ 0.50 & 80.70 $\pm$ 0.51 & 79.00 $\pm$ 0.47 & 42.10 $\pm$ 0.19 & 43.30 $\pm$ 0.19 & 42.10 $\pm$ 0.15 & 82.51 $\pm$ 0.18 & 77.00 $\pm$ 0.53 & 65.00 $\pm$ 0.43\\
        DCP            & \underline{96.67 $\pm$ 2.43} & 97.00 $\pm$ 2.00 & 96.60 $\pm$ 2.65 & \underline{73.43 $\pm$ 2.21} & 73.80 $\pm$ 2.40 & \underline{72.80 $\pm$ 2.48} & 92.76 $\pm$ 1.33 & 85.00 $\pm$ 1.29 & 83.60 $\pm$ 1.85 \\
        IMvGCN         & 83.10 $\pm$ 6.94 & 83.28 $\pm$ 6.36 & 79.64 $\pm$ 7.61& 29.48 $\pm$ 2.33 & 29.21 $\pm$ 3.75 & 25.68 $\pm$ 2.11  & 90.86 $\pm$ 0.81 & 83.51 $\pm$ 2.10 & 79.32 $\pm$ 1.69 \\
        IPMVSC         & 94.52 $\pm$ 3.38 & 94.59 $\pm$ 3.41 & 94.49 $\pm$ 3.03 & 60.55 $\pm$ 2.51 & 61.31 $\pm$ 2.51 & 61.39 $\pm$ 2.46& 88.05 $\pm$ 0.96 & 90.08 $\pm$ 2.12 & 79.11 $\pm$ 1.46  \\
        F\scalebox{0.85}{ACTOR}CL & 83.33 $\pm$ 3.72 & 83.54 $\pm$ 3.87 & 83.29 $\pm$ 4.02 & 32.14 $\pm$ 3.44 & 31.77 $\pm$ 3.44 & 29.6 $\pm$ 3.21 & 82.43 $\pm$ 1.84 & 82.45 $\pm$ 1.61 & 81.19 $\pm$ 1.83 \\
        CUMI           & 95.95 $\pm$ 1.86 & 96.88 $\pm$ 1.61 & 94.88 $\pm$ 2.33 & 71.60 $\pm$ 1.58 & 72.08 $\pm$ 1.74 & 70.51 $\pm$ 1.53  & 91.92 $\pm$ 0.31 & 83.93 $\pm$ 0.67 & 81.79 $\pm$ 1.26  \\
        DIMvLN         & 94.05 $\pm$ 3.73 & 95.31 $\pm$ 2.89 & 94.04 $\pm$ 3.74  & 71.00 $\pm$ 7.73 & \textbf{75.07 $\pm$ 5.32} & 70.59 $\pm$ 8.24& \underline{94.67 $\pm$ 0.42} & \underline{90.45 $\pm$ 1.59} & \underline{87.35 $\pm$ 1.04}\\
        \textbf{Ours}  & \textbf{100.00 $\pm$ 0.00} & \textbf{100.00 $\pm$ 0.00} & \textbf{100.00 $\pm$ 0.00}  & \textbf{74.05 $\pm$ 2.03} & \underline{74.70 $\pm$ 1.95} & \textbf{73.80 $\pm$ 1.89}& \textbf{94.98 $\pm$ 0.74} & 89.20 $\pm$ 1.60 & \textbf{87.70 $\pm$ 1.85}\\
    \hline
    \end{tabular}
    }
    \label{comparisionExp1}
\end{table*}

\begin{table*}
    \centering
    \caption{Classification results on the NUS, Scene-15, and NoisyMNIST datasets. Bold font denotes the best performance, and underline font indicates the second best performance.}
    \resizebox{\linewidth}{!}{
    \begin{tabular}{lccccccccc}
    \hline
        Datasets & \multicolumn{3}{c}{NUS} & \multicolumn{3}{c}{Scene-15} & \multicolumn{3}{c}{NoisyMNIST} \\
    \cmidrule(lr){2-4} \cmidrule(lr){5-7} \cmidrule(lr){8-10}
        Method & ACC (\%) & Precision (\%) & F1 (\%) & ACC (\%) & Precision (\%) & F1 (\%) & ACC (\%) & Precision (\%) & F1 (\%) \\
    \hline
        DCCA           & 36.73 $\pm$ 1.52 & 39.02 $\pm$ 1.93 & 36.66 $\pm$ 1.71 & 71.24 $\pm$ 1.50 & 70.99 $\pm$ 1.71 & 70.73 $\pm$ 1.55 & 92.86 $\pm$ 0.75 & 92.91 $\pm$ 0.73 & 92.87 $\pm$ 0.75  \\
        MIB            & 36.42 $\pm$ 1.05 & 36.51 $\pm$ 1.01 & 35.20 $\pm$ 1.08   & 66.42 $\pm$ 0.89 & 67.22 $\pm$ 1.00 & 65.25 $\pm$ 0.97 & 76.83 $\pm$ 0.66 & 77.03 $\pm$ 0.42 & 76.57 $\pm$ 0.71 \\
        TMC            & 42.38 $\pm$ 1.81 & 42.60 $\pm$ 3.53 & 37.70 $\pm$ 1.42   & 68.04 $\pm$ 1.03 & 72.40 $\pm$ 3.20 & 61.10 $\pm$ 0.94 & 93.00 $\pm$ 0.00 & 92.00 $\pm$ 0.00 & 92.00 $\pm$ 0.00 \\
        DUA-Nets       & 34.12 $\pm$ 3.19 & 39.19 $\pm$ 1.58 & 34.81 $\pm$ 1.23   & 57.80 $\pm$ 1.26 & 59.22 $\pm$ 1.59 & 54.30 $\pm$ 1.65 & 84.87 $\pm$ 0.50 & 85.22 $\pm$ 0.48 &  84.37 $\pm$ 0.56\\
        CMIB-Nets      & 36.31 $\pm$ 0.18 & 36.50 $\pm$ 0.19 & 36.10 $\pm$ 0.17   & 69.25 $\pm$ 0.16 & 68.40 $\pm$ 0.18 & 68.20 $\pm$ 0.15 & 93.76 $\pm$ 0.02 & 93.70 $\pm$ 0.05 & 93.70 $\pm$ 0.05\\
        DCP            & 39.71 $\pm$ 3.94 & 39.50 $\pm$ 3.50 & 38.50 $\pm$ 3.83 & \underline{75.01 $\pm$ 1.88} & \textbf{74.20 $\pm$ 1.72} & \textbf{73.60 $\pm$ 1.74} & 93.59 $\pm$ 1.46 & 93.70 $\pm$ 1.27 & 93.60 $\pm$ 1.56  \\
        IMvGCN         & \underline{43.60 $\pm$ 1.95} & 42.55 $\pm$ 2.03 & \underline{41.80 $\pm$ 1.96} & 55.83 $\pm$ 2.50 & 59.58 $\pm$ 3.64 & 51.01 $\pm$ 2.33 & 95.03 $\pm$ 0.22 & 94.98 $\pm$ 0.22 & 94.91 $\pm$ 0.22 \\
        IPMVSC         & 43.48 $\pm$ 1.78 & 42.44 $\pm$ 2.15 & 42.91 $\pm$ 2.07
        & 71.45 $\pm$ 1.07 & 72.45 $\pm$ 1.78 & 71.17 $\pm$ 1.30
        & 89.91 $\pm$ 0.32 & 90.25 $\pm$ 0.25 & 89.88 $\pm$ 0.26  \\
        F\scalebox{0.85}{ACTOR}CL & 30.63 $\pm$ 1.40 & 32.22 $\pm$ 3.37 & 28.29 $\pm$ 1.30
        & 41.58 $\pm$ 2.40 & 42.41 $\pm$ 2.94 & 39.47 $\pm$ 2.49
        & 93.56 $\pm$ 0.96 & 93.61 $\pm$ 0.96 & 93.57 $\pm$ 0.96 \\
        CUMI           & 42.42 $\pm$ 1.33 & \underline{43.47 $\pm$ 2.46} & 41.51 $\pm$ 1.68& 73.19 $\pm$ 1.52 & 72.71 $\pm$ 2.13 & 71.00 $\pm$ 1.81 & 95.48 $\pm$ 0.18 & 95.46 $\pm$ 0.19 & 95.44 $\pm$ 0.20 \\
        DIMvLN         & 31.81 $\pm$ 4.95 & 41.19 $\pm$ 5.52 & 30.52 $\pm$ 6.17 & 65.88 $\pm$ 5.55 & 70.00 $\pm$ 3.20 & 64.42 $\pm$ 5.72   & \textbf{95.70 $\pm$ 0.19} & \underline{95.68 $\pm$ 0.18} & \underline{95.64 $\pm$ 0.19}\\
        \textbf{Ours}  & \textbf{44.65 $\pm$ 1.05} & \textbf{46.40 $\pm$ 1.50} & \textbf{44.90 $\pm$ 1.14}  & \textbf{75.45 $\pm$ 1.05} & \underline{73.50 $\pm$ 2.46} & \underline{72.00 $\pm$ 1.48}  
        & \underline{95.62 $\pm$ 0.76} & \textbf{95.70 $\pm$ 0.78} & \textbf{95.70 $\pm$ 0.78} \\
    \hline
    \end{tabular}
    }
    \label{comparisionExp2}
\end{table*}

\subsubsection{Baselines}

We compare our model with several state-of-art baselines. \textbf{DCCA} \cite{andrew2013deep} extracts common information by using neural network to perform nonlinear transformation on each view. \textbf{MIB} \cite{federici2020learning} learns robust representations by maximizing mutual information between views while minimizing superfluous information. \textbf{TMC} \cite{han2022trusted} formulates a reliable multi-view classification framework by introducing the uncertainty estimation theory. \textbf{DUA-Nets} \cite{geng2021uncertainty} dynamically assigns weights to individual views of different samples by estimating the uncertainty of the data. \textbf{CMIB-Nets} \cite{wan2021multi} integrates shared and view-specific representations while discarding superfluous information using the information bottleneck. \textbf{DCP} \cite{lin2022dual} establishes an information theoretic framework that allows for consistency learning and data recovery. \textbf{IMvGCN} \cite{wu2023interpretable} addresses the multi-view classification challenge in scenarios with limited labeled samples by integrating reconstruction error and Laplace embedding. \textbf{IPMVSC} \cite{hu2022multi} learns the landmarks of each view to represent the potential data distribution. \textbf{F\scalebox{0.85}{ACTOR}CL} \cite{liang2023factorized} factorizes task-relevant information into shared and unique representations. \textbf{CUMI} \cite{zhang2024discovering} exploits the common information among multi-view data for classification. \textbf{DIMvLN} \cite{jiang2024deep} recovers the semantic information of unlabeled multi-view data by designing a pseudo-label generation strategy.

\subsubsection{Implementation Details}

For a fair comparison, we tune the parameters of the existing methods based on the recommended settings provided in the corresponding literature. For our method, $\beta_1$ and $\beta_2$ are fixed at the constant value of $10^{-4}$, while $\beta_3$ and $\beta_4$ are tuned over the sets $\{10^{0}, 10^{1}, 10^{2}, 10^{3}\}$ and $\{10^{-3}, 10^{-2}, 10^{-1}, 10^{0}\}$, respectively, to achieve optimal performance. We evaluate the model performance using Accuracy (ACC), Precision, and F-score (F1). All results are recorded by averaging over 10 trials.

\subsection{Experimental Results}

\subsubsection{Performance Comparison with Baselines}

Tables. \ref{comparisionExp1} and \ref{comparisionExp2} show the classification results of various methods on six datasets. It is seen that our proposed CIML model almost achieves the best ACC on all tested datasets. Apart from the ACC metric, CIML also outperforms other  baselines or nearly matches the best performance in terms of Precision and F1 score, with only a slight inferiority. This advanced performance achieved by our model can be attributed to its formulation of a comprehensive learning framework based on information theory, which ensures the completeness in learning both common and unique representations. In addition, we observe that as a representative trustworthy learning method, TMC maintains competitive performance among exiting methods, thanks to its capability of measuring data quality to generate reliable results. Another remarkable baseline method is DCP, which ranks second in ACC on several datasets. This is because DCP has the potential to maximize cross-view consistency through contrastive learning. Compared with CUMI, which relies solely on common information derived from information theory for downstream tasks, our model achieves consistently better performance. This highlights the advantage of employing a comprehensive framework that jointly learns both common and unique representations. In summary, while several representative baseline models exhibits competitive results, our model demonstrates clear overall superiority. 

\begin{table*}[!htb]
    \centering
    \tiny
    \caption{Ablation study on six datasets. Bold font indicates the best performance.}
    \resizebox{\linewidth}{!}{
    \begin{tabular}{lcccccc}
    \hline
        Method    & MSRC-v1 & LandUse-21 & Caltech101-20 & NUS   & Scene-15 & NoisyMNIST\\
    \hline
    	CIML-v1 & 98.10 $\pm$ 0.95 & 65.90 $\pm$ 1.44 & 93.64 $\pm$ 0.82 & 35.00 $\pm$ 1.65 & 70.88 $\pm$ 1.73 & 94.41 $\pm$ 0.60 \\
        CIML-v2 & 97.14 $\pm$ 0.95 & 63.90 $\pm$ 1.72 & 93.56 $\pm$ 0.93 & 37.71 $\pm$ 2.45 & 72.24 $\pm$ 1.59 & 93.18 $\pm$ 2.87 \\
        CIML      & \textbf{100.00 $\pm$ 0.00} & \textbf{74.05 $\pm$ 2.03} & \textbf{94.98 $\pm$ 0.74} & \textbf{44.65 $\pm$ 1.05} & \textbf{75.45 $\pm$ 1.05} & \textbf{95.62 $\pm$ 0.76} \\
    \hline
    \end{tabular}
    }
    \label{tab:ablation experiment}
\end{table*}

%
%
%

\subsubsection{Ablation Study}

We conduct ablation experiments to investigate the contributions of learned common representation and unique representations. Specifically, in the first test, we remove the loss term of $\mathcal{L}_c$, while in the second test, we remove the loss term of $\mathcal{L}_u$. The corresponding variants are named CIML-v1 and CIML-v2, respectively. The experimental results are shown in Table \ref{tab:ablation experiment}. It is seen that CIML outperforms its two variants on all tested datasets, implying that both common and unique representations contribute to the model's performance. In addition, we find that common representation does not always provide more predictive information than the unique representation. For example, on the LandUse-21 dataset, CIML-v1 that maintains the learning module of unique representation outperforms CIML-v2 that preserves the learning module of common representation. This further emphasizes the importance of learning unique representation.

\subsubsection{Parameter and Dimensionality Sensitivity Analysis}

%


\begin{figure*}[htbp]
	\centering
	\includegraphics[scale=0.65]{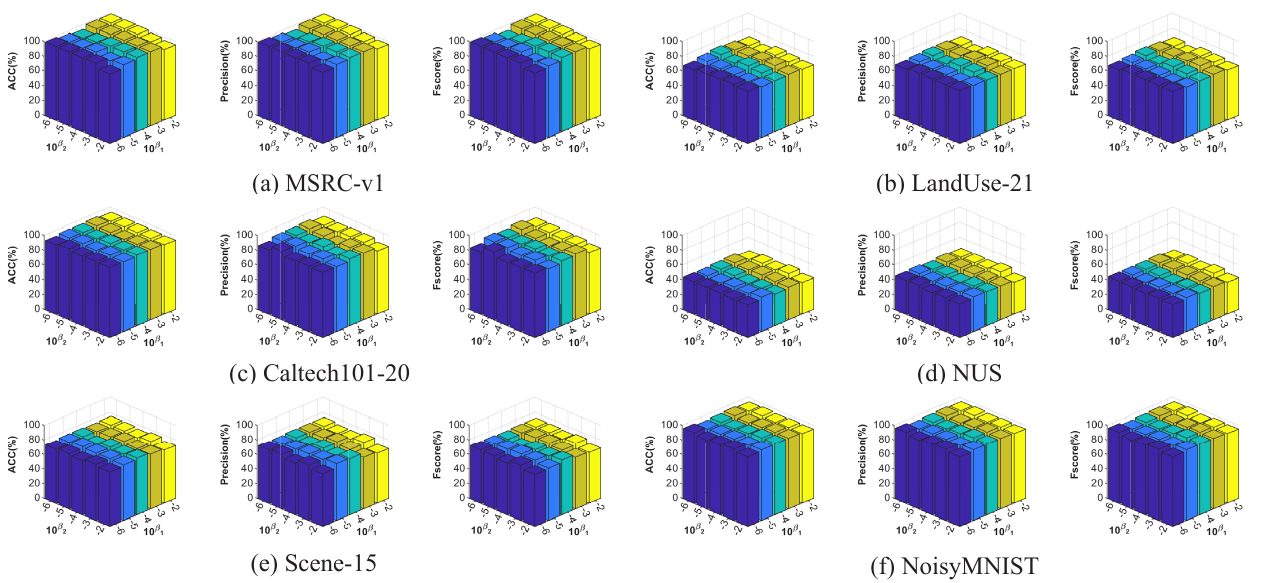} 
	\caption{Parameter sensitivity analysis with respect to $\beta_1$ and $\beta_2$}
	\label{beta12}
\end{figure*}

\begin{figure*}[htbp]
	\centering
	\includegraphics[scale=0.65]{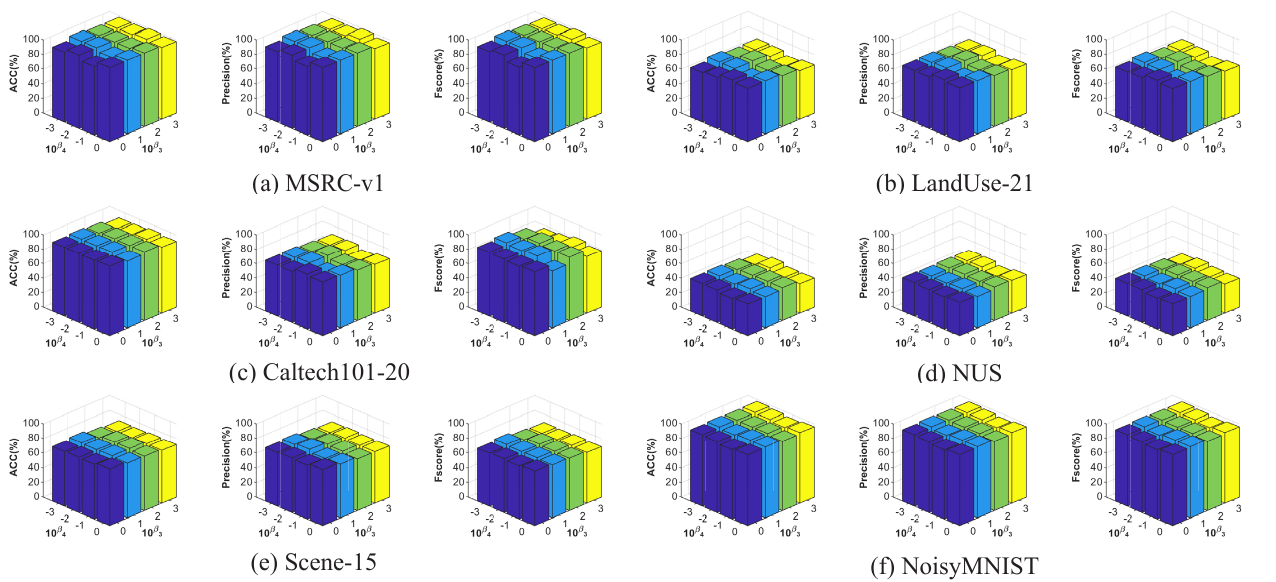} 
	\caption{Parameter sensitivity analysis with respect to $\beta_3$ and $\beta_4 $}
	\label{beta34}
\end{figure*}

\begin{figure*}[htbp]
	\centering
	\includegraphics[scale=0.75]{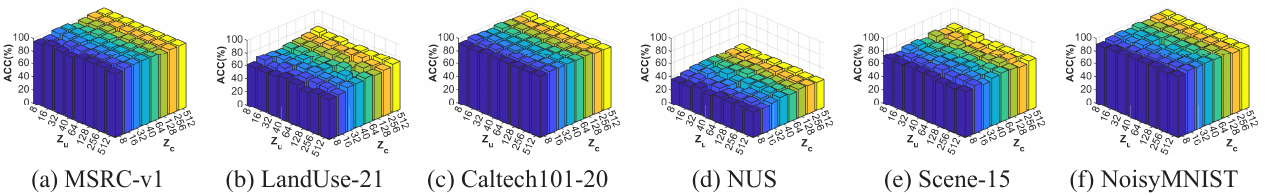} 
	\caption{Dimensionality sensitivity analysis with respect to $\boldsymbol{Z}_c$ and $\boldsymbol{Z}_u$}
	\label{fig:dim_sen}
\end{figure*}

We investigate the sensitivity of the hyperparameters $\beta_1$ and $\beta_2$, as well as $\beta_3$ and $\beta_4$, as illustrated in Figs. \ref{beta12} and \ref{beta34}. For $\beta_1$ and $\beta_2$, we constrain their values to the range $\{10^{-6}, \ldots, 10^{-2}\}$ and observe that variations within this interval have negligible impact on model performance. This finding indicates that the model is not sensitive to the choice of $\beta_1$ and $\beta_2$, which is consistent with the common practice of assigning small values to regularization parameters that govern information compression in the IB principle. For $\beta_3$ and $\beta_4$, we observe that the proposed model maintains stable performance when $\beta_3$ is set within the range $\{10^0, \ldots, 10^3\}$ and $\beta_4$ within $\{10^{-3}, \ldots, 10^0\}$. This observation aligns with the intuitive expectation that the balance parameter associated with learning common representations should be relatively large, whereas the parameter governing unique representations should be relatively small. Therefore, the model can be expected to maintain satisfactory performance even when the parameters are chosen at random. We also investigate the dimensionality sensitivity of $\boldsymbol{Z}_c$ and $\boldsymbol{Z}_u$, as shown in Fig. \ref{fig:dim_sen}. We see a slight but not very evident increase in model performance with the increase of dimensions $\boldsymbol{Z}_c$ and $\boldsymbol{Z}_u$. A moderate dimension can be determined under a balance between performance and efficiency.

\subsubsection{Convergence Analysis and Visualization}

\begin{figure*}[htbp]
	\centering
	\includegraphics[scale=0.75]{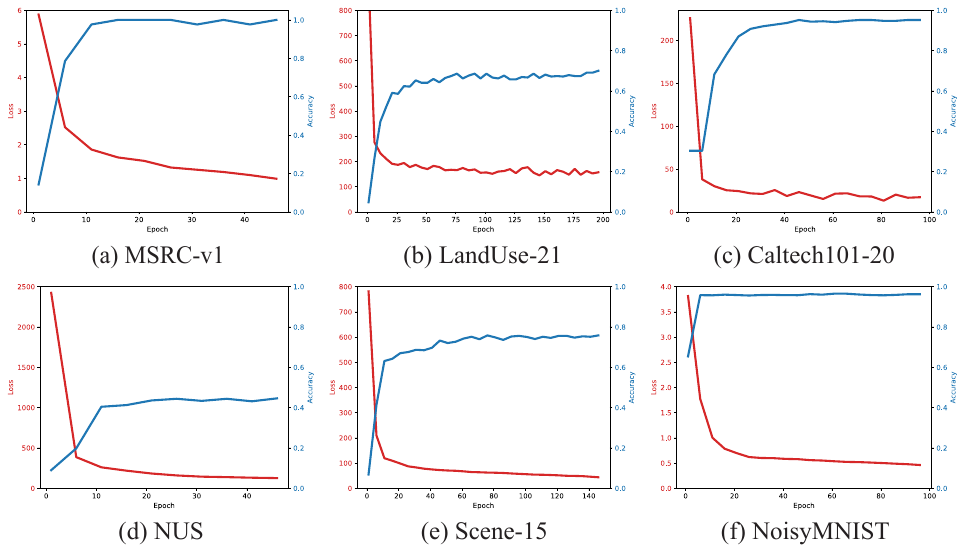} 
	\caption{Evolutionary curves of training loss and accuracy with respect to epoch.}
	\label{fig:convergence}
\end{figure*}

%
%
%

\begin{figure*}  
\centering  
  
\begin{subfigure}{.28\textwidth}  
  \centering  
  \includegraphics[width=\linewidth]{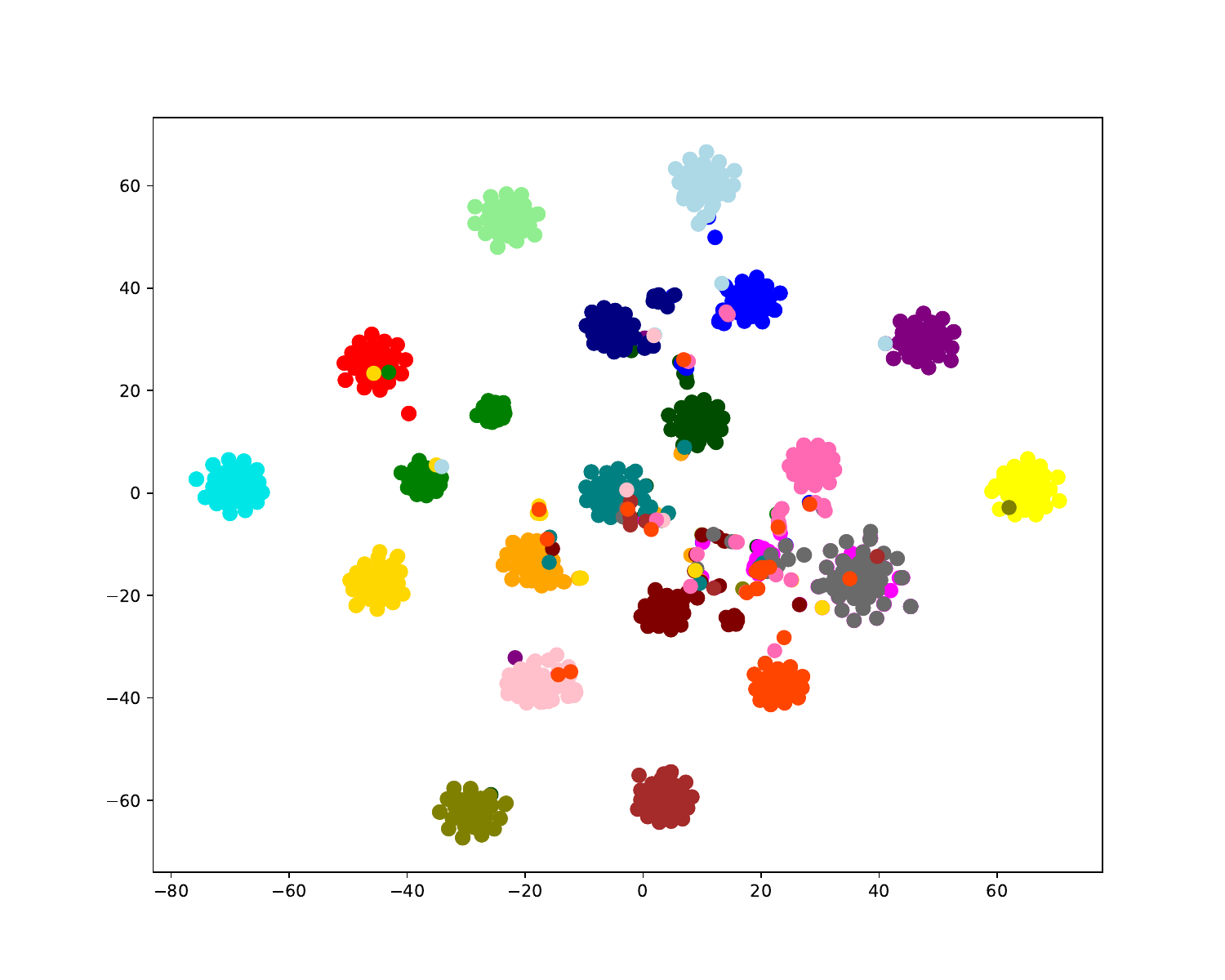}  
  \caption{LandUse-21}  
\end{subfigure}%
\begin{subfigure}{.28\textwidth}  
  \centering  
  \includegraphics[width=\linewidth]{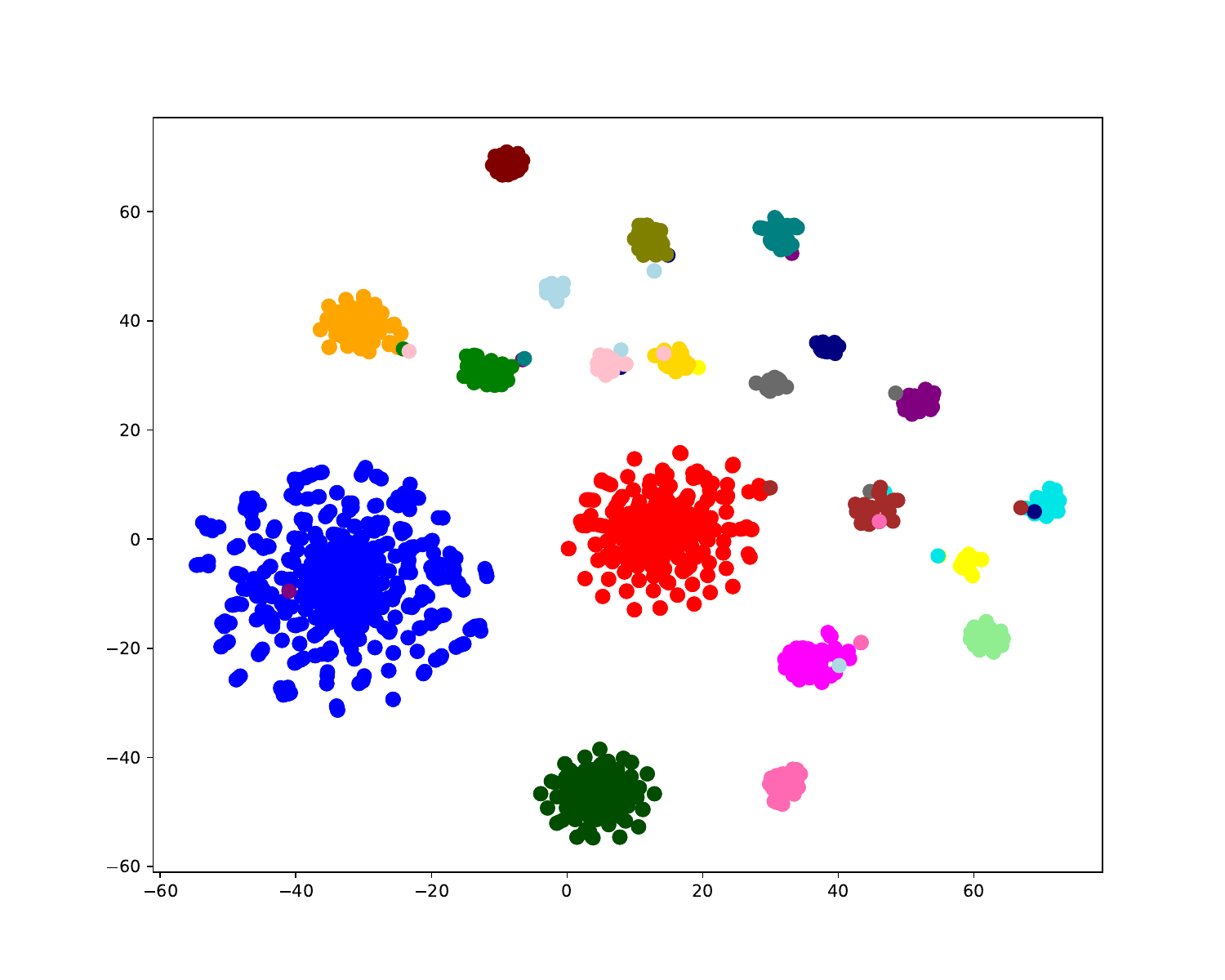}  
  \caption{Caltech101-20}  
\end{subfigure}  
\begin{subfigure}{.28\textwidth}  
  \centering  
  \includegraphics[width=\linewidth]{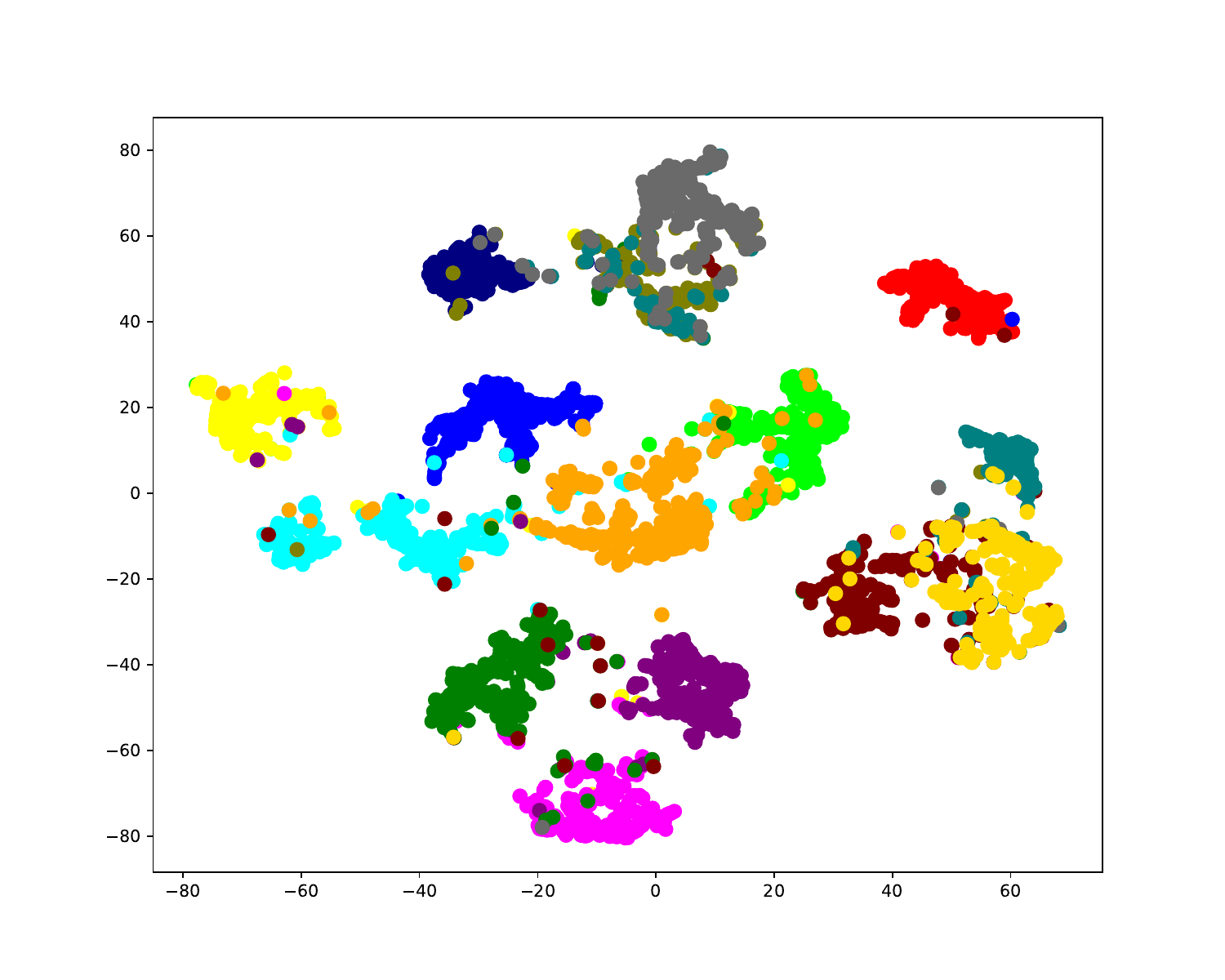}  
  \caption{Scene15}  
\end{subfigure}%
  
\caption{T-SNE visualization of our model on the LandUse-21, Caltech101-20, and Scene15 datasets.}
\label{fig: t-SNE}  
\end{figure*} 

Fig. \ref{fig:convergence} shows the evolutionary curves of training loss and accuracy with respect to epoch on different datasets. It is observed that the training loss of our model decreases with epoch, and the model nearly approaches to a steady state when the epoch exceeds 40 on all tested datasets, while the accuracy improves as the epochs increase. This implies that the training of our proposed CIML model is convergent, and the model parameters will arrive at an optimal state within a certain number of epochs. To provide a more intuitive visualization, we present the t-Distributed Stochastic Neighbor Embedding (T-SNE) \cite{van2008visualizing} results of CIML on three datasets, as shown in Fig. \ref{fig: t-SNE}. It is observed that our model achieves clear classification clusters, further demonstrating its superiority in distinguishing different classes.

%

\section{Conclusions and Future Work}

In this paper, we propose a comprehensive information-theoretic multi-view learning framework called CIML, which considers the predictive capabilities of both common and unique information. For common representation learning, CIML leverages G$\acute{a}$cs-K$\mathaccent"707F o$rner common information to extract the shared information across different views. This approach overcomes the limitation of earlier methods, which aggregate view-pair common information, a strategy that does not truly capture the common information. CIML then employs IB to compress the learned common information, retaining only the task-relevant features. For unique representation learning, CIML utilizes IB to obtain compressed unique representations while simultaneously minimizing mutual information between unique and common representations, as well as among different unique representations. This ensures that the learned unique representation for each view remains highly distinct. Upon combining the learned common and unique representations, a comprehensive representation is obtained, which can be effectively utilized for downstream tasks. Extensive experimental results demonstrate that CIML outperforms recent state-of-the-art baselines.

Despite we have proposed a comprehensive information-theoretic framework in this work, CIML has a notable limitation in handling incomplete multi-view problem. It cannot directly process situations where multi-view data is incomplete. However, the modular and flexible nature of the CIML framework provides a natural pathway for extension to incomplete multi-view settings. As part of future work, we plan to extend CIML into a robust framework capable of jointly learning representations and recovering missing views through a unified information-theoretic objective.
\appendices

\section{Details on deriving Eq. (9)}

According to the definition of mutual information, $I(\boldsymbol{Z}_c;\boldsymbol{C})$ can be calculated by
\begin{equation}
\begin{aligned}
I(\boldsymbol{Z}_c;\boldsymbol{C}) &= \iint p(\boldsymbol{z}_c,\boldsymbol{c}) \log \frac{p(\boldsymbol{z}_c,\boldsymbol{c})}{p(\boldsymbol{z}_c)p(\boldsymbol{c})} \, d\boldsymbol{c}\,d\boldsymbol{z}_c \\
         &= \iint p(\boldsymbol{z}_c,\boldsymbol{c}) \log \frac{p(\boldsymbol{z}_c|\boldsymbol{c})}{p(\boldsymbol{z}_c)} \, d\boldsymbol{c}\,d\boldsymbol{z}_c. \\
\end{aligned}
\tag{A1}
\end{equation}

Let $r(\boldsymbol{z}_c)$ be a variational approximation to $p(\boldsymbol{z}_c)$, and introduce the definition of Kullback-Leibler (KL) divergence between $r(\boldsymbol{z}_c)$ and $p(\boldsymbol{z}_c)$
\begin{equation}
D_{KL}(p(\boldsymbol{z}_c) \parallel r(\boldsymbol{z}_c)) = \iint p(\boldsymbol{z}_c) \log \frac{p(\boldsymbol{z}_c)}{r(\boldsymbol{z}_c)} \, d\boldsymbol{c}\,d\boldsymbol{z}_c.
\tag{A2}
\end{equation}

Since $D_{KL}(p(\boldsymbol{z}_c) \parallel r(\boldsymbol{z}_c)) \geq 0$, from Eq. (A1) we easily  have

\begin{equation}
\begin{aligned}
\iint p(\boldsymbol{z}_c,\boldsymbol{c}) &\log \frac{p(\boldsymbol{z}_c|\boldsymbol{c})}{p(\boldsymbol{z}_c)} \, d\boldsymbol{c}\,d\boldsymbol{z}_c \\
&\leq \iint p(\boldsymbol{z}_c,\boldsymbol{c}) \log \frac{p(\boldsymbol{z}_c|\boldsymbol{c})}{r(\boldsymbol{z}_c)} \, d\boldsymbol{c}\,d\boldsymbol{z}_c.
\end{aligned}
\tag{A3}
\end{equation}

By combing Eqs. (A1) and (A3), we derive
\begin{equation}
\begin{aligned}
I(\boldsymbol{Z}_c;\boldsymbol{C}) &= \iint p(\boldsymbol{z}_c,\boldsymbol{c}) \log \frac{p(\boldsymbol{z}_c|\boldsymbol{c})}{p(\boldsymbol{z}_c)} \, d\boldsymbol{c}\,d\boldsymbol{z}_c \\
         &\leq \iint p(\boldsymbol{z}_c,\boldsymbol{c}) \log \frac{p(\boldsymbol{z}_c|\boldsymbol{c})}{r(\boldsymbol{z}_c)} \, d\boldsymbol{c}\,d\boldsymbol{z}_c \\
         &= \iint p(\boldsymbol{c})p(\boldsymbol{z}_c|\boldsymbol{c}) \log \frac{p(\boldsymbol{z}_c|\boldsymbol{c})}{r(\boldsymbol{z}_c)} \, d\boldsymbol{c}\,d\boldsymbol{z}_c. \\
\end{aligned}
\tag{A4}
\end{equation}

In practice, we can approximate the double integration over $\boldsymbol{z}_c$ and $\boldsymbol{c}$ by Monte Carlo sampling, allowing us to obtain an empirically computable upper bound: 
\begin{equation}
\begin{aligned}
I(\boldsymbol{Z}_c;\boldsymbol{C}) &\approx \frac{1}{N} \sum^N_{k=1}{\int p(\boldsymbol{z}_c|\boldsymbol{c}_k) \log \frac{p(\boldsymbol{z}_c|\boldsymbol{c}_k)}{r(\boldsymbol{z}_c)}} \, d\boldsymbol{z}_c \\
         &= \frac{1}{N} \sum^N_{k=1}{D_{KL}(p(\boldsymbol{z}_c|\boldsymbol{c}_k) \parallel r(\boldsymbol{z}_c))}. \\
\end{aligned}
\tag{A5}
\end{equation}

We assume that $r(\boldsymbol{z}_c)$ is normally distributed, and use reparameterization track to produce samples of $\boldsymbol{z}_{c,k}=\boldsymbol{z}_c(\boldsymbol{c}_k)=\mu_{\boldsymbol{z}_c}(\boldsymbol{c}_k)+\sigma_{\boldsymbol{z}_c}(\boldsymbol{c}_k) \eta_k$ from $p(\boldsymbol{z}_c|\boldsymbol{c}_k)$, where $\eta_k$ is drawn from a standard normal distribution. Then the above formula can be further written into
\begin{equation}
\begin{aligned}
I(\boldsymbol{Z}_c;\boldsymbol{C}) \approx &\frac{1}{2N} \sum^N_{k=1}[Tr(\sigma_{\boldsymbol{z}_c}(\boldsymbol{c}_k)+\rVert \mu_{\boldsymbol{z}_c}(\boldsymbol{c}_k) \lVert^2\\
&-d_{\boldsymbol{z}_c}-ln\,det\sigma_{\boldsymbol{z}_c}(\boldsymbol{c}_k)]. 
\end{aligned}
\tag{A6}
\end{equation}

\section{Details on deriving Eq. (19)}

Firstly, let's review the definition of completeness of representation:

\textbf{Definition}: The joint representation $\boldsymbol{Z} = (\boldsymbol{Z}_c, \boldsymbol{Z}^{(1)}_u, \dots, \boldsymbol{Z}^{(v)}_u)$ is predictively sufficient for the target $\boldsymbol{Y}$ if, given that the original views $\boldsymbol{X}=(\boldsymbol{X}^{(1)}, \cdots, \boldsymbol{X}^{(v)})$ fully determine $\boldsymbol{Y}$, the joint representation $\boldsymbol{Z}$ approximately determines $\boldsymbol{Y}$.  Mathematically, we need:
\begin{equation}
	H(\boldsymbol{Y}|\boldsymbol{Z}_c, \boldsymbol{Z}^{(1)}_u, \dots, \boldsymbol{Z}^{(v)}_u) \approx H(\boldsymbol{Y}|\boldsymbol{X}^{(1)}, \dots, \boldsymbol{X}^{(v)}). 
	\tag{B1}
\end{equation}

Here is the proof:

The conditional entropy $H(\boldsymbol{Y}|\boldsymbol{Z}_c, \boldsymbol{Z}^{(1)}_u, \dots, \boldsymbol{Z}^{(v)}_u)$ quantifies the average uncertainty about $\boldsymbol{Y}$ when $\boldsymbol{Z}_c, \boldsymbol{Z}^{(1)}_u, \dots, \boldsymbol{Z}^{(v)}_u$ are known. It can be equivalently expressed in terms of mutual information as follows:
\begin{equation}
\begin{aligned}
	H(\boldsymbol{Y}|\boldsymbol{Z}_c, \boldsymbol{Z}^{(1)}_u,& \dots, \boldsymbol{Z}^{(v)}_u)\\
	&= H(\boldsymbol{Y}) - I(\boldsymbol{Z}_c, \boldsymbol{Z}^{(1)}_u, \dots, \boldsymbol{Z}^{(v)}_u; \boldsymbol{Y}).
\end{aligned}
	\tag{B2}
\end{equation}
By using the chain rule, the mutual information term in Eq. (B2) can be decomposed as:
\begin{equation}
\begin{aligned}
	I(\boldsymbol{Z}_c, \boldsymbol{Z}^{(1)}_u,& \dots, \boldsymbol{Z}^{(v)}_u; \boldsymbol{Y})\\
	&= I(\boldsymbol{Z}_c;\boldsymbol{Y}) + I(\boldsymbol{Z}^{(1)}_u, \dots, \boldsymbol{Z}^{(v)}_u;\boldsymbol{Y}|\boldsymbol{Z}_c).
\end{aligned}
	\tag{B3}
\end{equation} 

In the learning process, we enforce constraints to ensure that the unique components $\boldsymbol{Z}^{(i)}_u$ are both independent of the common component $\boldsymbol{Z}_c$ and mutually independent across views, i.e., $min \;\; \sum_{i=1}^v \sum_{j=1,j\neq i}^{v}{I(\boldsymbol{Z}^{(i)}_u;\boldsymbol{Z}^{(j)}_u)}$ and  $min \;\; \sum_{i=1}^{v} I(\boldsymbol{Z}^{(i)}_u; \boldsymbol{Z}_c)$. Under ideal conditions, these constraints yield $I(\boldsymbol{Z}^{(i)}_u;\boldsymbol{Z}_c)\approx 0$ and $I(\boldsymbol{Z}^{(i)}_u;\boldsymbol{Z}^{(j)}_u)\approx 0$, implying that $\boldsymbol{Z}_c$ and all $\boldsymbol{Z}^{(i)}_u$ are mutually independent. Consequently, the conditional mutual information in Eq. (B3) simplifies as:
\begin{equation}
\begin{aligned}
	I(\boldsymbol{Z}^{(1)}_u,\dots,\boldsymbol{Z}^{(v)}_v;\boldsymbol{Y}|\boldsymbol{Z}_c) \approx& \sum^v_{i=1} I(\boldsymbol{Z}^{(i)}_u;\boldsymbol{Y}|\boldsymbol{Z}_c)\\
	\approx& I(\boldsymbol{Z}^{(i)}_u;\boldsymbol{Y}).
\end{aligned}
	\tag{B4}
\end{equation}
Substituting Eq. (B4) into Eq. (B3), we obtain
\begin{equation}
	I(\boldsymbol{Z}_c, \boldsymbol{Z}^{(1)}_u, \dots, \boldsymbol{Z}^{(v)}_u; \boldsymbol{Y}) \approx I(\boldsymbol{Z}_c;\boldsymbol{Y}) + \sum^v_{i=1}I(\boldsymbol{Z}^{(i)}_u;\boldsymbol{Y}).
	\tag{B5}
\end{equation} 

Now consider the learning objective for the common representation $\boldsymbol{Z}_c$, which involves learning the the shared information across all views via GK while  performing information compression with IB, as shown in Eqs. (1) and (2). It worth noting that the second term of Eq. (2) serves as a penalty term. When $\beta_1$ is small, this penalty is weak, allowing $\boldsymbol{Z}_c$ to preserve most of the predictive information in $\boldsymbol{C}$ relevant to $\boldsymbol{Y}$, such that  
\begin{equation}
	I(\boldsymbol{Z}_c;\boldsymbol{Y}) \approx I(\boldsymbol{C};\boldsymbol{Y}).
	\tag{B6}
\end{equation}  
Since $\boldsymbol{C}$ aggregates information across all views and $\beta_1$ is typically chosen to be small in practice, $\boldsymbol{Z}_c$ effectively captures the shared predictive information relevant to $\boldsymbol{Y}$.

Similarly, for the unique representations $\boldsymbol{Z}^{(i)}_u$, as governed by Eq. (B4)-(B6), the learning process ensures that $\sum^{v}_{i=1}I(\boldsymbol{Z}^{(i)}_u;\boldsymbol{Y})$ approximates the total unique information with small penalty on $I(\boldsymbol{Z}^{(i)}_u;\boldsymbol{X}^{(i)})$ and independence constraints. 

Therefore, under practical settings with small $\beta_1$ and $\beta_2$, the combined mutual information between $\boldsymbol{Z}_c, \boldsymbol{Z}^{(1)}_u, \dots, \boldsymbol{Z}^{(v)}_u$ and $\boldsymbol{Y}$ closely matches that of the original views
\begin{equation}
	I(\boldsymbol{Z}_c;\boldsymbol{Y}) + \sum^v_{i=1}I(\boldsymbol{Z}^{(i)}_u;\boldsymbol{Y}) \approx I(\boldsymbol{X}^{(1)},\dots,\boldsymbol{X}^{(v)};\boldsymbol{Y}).
	\tag{B7}	
\end{equation}   

Substituting Eq. (B7) into Eq. (B5), and then into Eq. (B2), we obtain
\begin{equation}
\begin{aligned}
	H(\boldsymbol{Y}) - I(\boldsymbol{Z}_c, \boldsymbol{Z}^{(1)}_u,& \dots, \boldsymbol{Z}^{(v)}_u; \boldsymbol{Y})\\
	&\approx H(\boldsymbol{Y}) - I(\boldsymbol{X}^{(1)}, \dots, \boldsymbol{X}^{(v)}; \boldsymbol{Y}).
\end{aligned}
	\tag{B8}
\end{equation} 
Consequently, $H(\boldsymbol{Y}|\boldsymbol{Z}_c, \boldsymbol{Z}^{(1)}_u, \dots, \boldsymbol{Z}^{(v)}_u) \approx H(\boldsymbol{Y}|\boldsymbol{X}^{(1)}, \dots, \boldsymbol{X}^{(v)})$ holds. This completes the proof.


\ifCLASSOPTIONcaptionsoff
  \newpage
\fi

\bibliographystyle{IEEEtran}
\bibliography{IEEEabrv,IEEEexample}

\end{document}